\documentclass[runningheads]{llncs}
\usepackage{graphicx}
\usepackage{epstopdf}
\usepackage{comment}
\usepackage{amsmath,amssymb} 
\usepackage{color}
\usepackage{epsfig}
\usepackage{acro}
\usepackage{xspace}
\usepackage{multirow}
\usepackage{booktabs}
\usepackage{makecell}
\usepackage{amssymb}
\usepackage{marvosym}
\usepackage{hyperref}

\acsetup{first-style=short}

\makeatletter
\DeclareRobustCommand\onedot{\futurelet\@let@token\@onedot}
\def\@onedot{\ifx\@let@token.\else.\null\fi\xspace}

\def\eg{\emph{e.g}\onedot} 
\def\ie{\emph{i.e}\onedot}

\def\etal{\emph{et al}\onedot}
\makeatother

\DeclareAcronym{CAD}{
short=CAD,
long=computer-aided diagnosis,
}

\DeclareAcronym{AASN}{
short=AASN,
long=anatomy-aware Siamese network
}

\DeclareAcronym{PXR}{
short=PXR,
long=pelvic X-ray
}

\DeclareAcronym{CXR}{
short=CXR,
long=chest X-ray
}

\DeclareAcronym{TPS}{
short=TPS,
long=thin-plate spline
}

\DeclareAcronym{GCN}{
short=GCN,
long=graph convolutional network
}

\DeclareAcronym{ROI}{
short=ROI,
long=region of interest
}

\DeclareAcronym{BCE}{
short=BCE,
long=binary cross entropy
}

\makeatletter
\newcommand{\printfnsymbol}[1]{%
  \textsuperscript{\@fnsymbol{#1}}%
}
\makeatother

\begin{document}
\pagestyle{headings}
\mainmatter
\def\ECCVSubNumber{4300}  

\title{Anatomy-Aware Siamese Network: Exploiting Semantic Asymmetry for Accurate Pelvic Fracture Detection in X-ray Images} 

\titlerunning{Anatomy-Aware Siamese Network for Pelvic Fracture Detection}
%
\author{Haomin Chen\thanks{equal contribution}\inst{1,2} \and
Yirui Wang\printfnsymbol{1}\inst{1} \and
Kang Zheng\inst{1}  \and
Weijian Li\inst{3}  \and
Chi-Tung Chang\inst{5}  \and
Adam P. Harrison\inst{1}  \and
Jing Xiao\inst{4}  \and
Gregory D. Hager\inst{2}  \and
Le Lu\inst{1}  \and
Chien-Hung Liao\inst{5}  \and
Shun Miao\inst{1}}

\authorrunning{Chen et al.}
%
\institute{PAII Inc., Bethesda, MD, USA \and
Departemnt of Computer Science, Johns Hopkins University, Baltimore, MD, USA \and
Department of Computer Science, University of Rochester, NY, USA \and
Ping An Technology, Shenzhen, China \and
Chang Gung Memorial Hospital, Linkou, Taiwan, ROC}
\maketitle

\begin{abstract}
Visual cues of enforcing bilaterally symmetric anatomies as normal findings are widely used in clinical practice to disambiguate subtle abnormalities from medical images. So far, inadequate research attention has been received on effectively emulating this practice in \ac{CAD} methods.
In this work, we exploit semantic anatomical symmetry or asymmetry analysis in a complex \ac{CAD} scenario, i.e., anterior pelvic fracture detection in trauma \acp{PXR}, where semantically pathological (refer to as fracture) and non-pathological (\eg pose) asymmetries both occur. Visually subtle yet pathologically critical fracture sites can be missed even by experienced clinicians, when limited diagnosis time is permitted in emergency care. We propose a novel fracture detection framework that builds upon a Siamese network enhanced with a spatial transformer layer to holistically analyze symmetric image features. Image features are spatially formatted to encode bilaterally symmetric anatomies. A new contrastive feature learning component in our Siamese network is designed to optimize the deep image features being more salient corresponding to the underlying semantic asymmetries (caused by pelvic fracture occurrences). Our proposed method have been extensively evaluated on 2,359 \acp{PXR} from unique patients (the largest study to-date), and report an area under ROC curve score of 0.9771. This is the highest among state-of-the-art fracture detection methods, with improved clinical indications.
\end{abstract}

\section{Introduction}
\label{sec:introduction}



\begin{figure}[t]
   \begin{center}
      \includegraphics[width=0.7\linewidth]{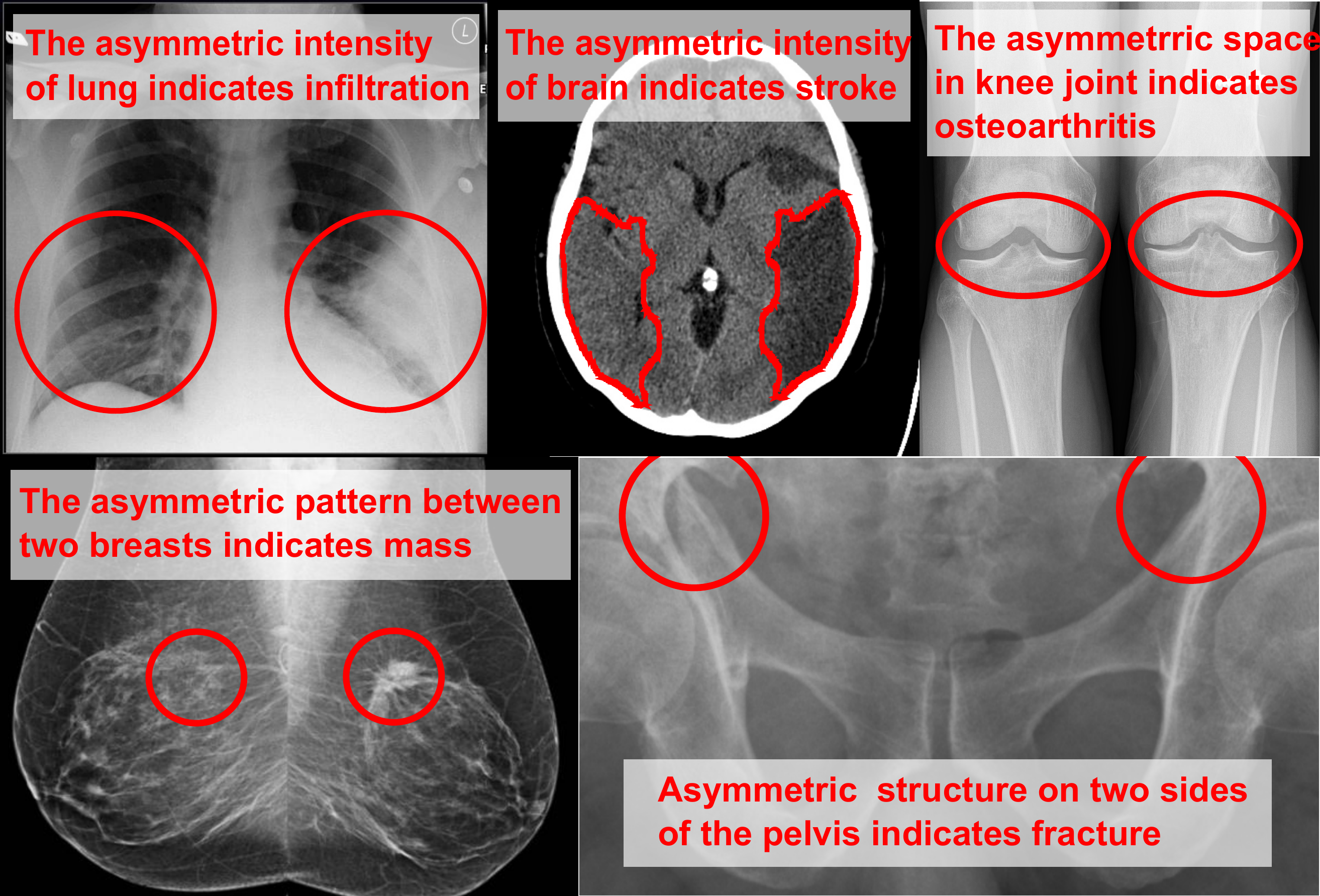}
   \end{center}
   \caption{Example medical images where anatomical symmetry helps to detect abnormalities. The top 3 images represents infiltration in chest X-Rays, stroke in brain CT, and osteoarthritis in knee X-Rays. The bottom 2 images represent masses in mammography and fractures in PXRs. These abnormalities can be better differentiated when the anatomically symmetric body parts are compared.}
   \label{fig:intro} 
\end{figure}

\acresetall
The \ac{CAD} of abnormalities in medical images is among the most promising applications of computer vision in healthcare. In particular, X-ray \ac{CAD} represents an important research focus~\cite{haomin_hierarchy,chestxray8,chexnet,li_thoracic_2018,CheXpert,PadChest,lu2020learning}. However, the high variations of abnormalities in medical imagery pose non-trivial challenges in differentiating pathological abnormalities from radiological patterns caused by normal anatomical and imaging-condition differences. At the same time, many anatomical structures are bilaterally symmetric (e.g., the brain, skeleton and breast) which suggests that the detection of abnormal radiological findings can exploit semantically symmetric anatomical regions (Figure~\ref{fig:intro}). Indeed, using bilaterally symmetric visual cues to confirm suspicious findings is a strongly recommended and widely adopted clinical practice~\cite{clohisy2008systematic}. Our aim is to emulate this practice in \ac{CAD} and apply it to the problem of effectively detecting subtle but critical anterior pelvic fractures in trauma \acp{PXR}.

Several studies have investigated the use of symmetry cues for \ac{CAD}, aiming to find abnormalities in brain structures in neuro-imaging~\cite{wachinger2015brainprint,konukoglu2012wesd,Alzheimer}, breasts in mammograms~\cite{mammo}, and stroke in CT~\cite{stroke}. All of these works directly employ symmetry defined on the image or shape space. However, under less constrained scenarios, especially the ones using projection-based imaging modalities in an emergency room setting, \eg, \acp{PXR}, image asymmetries do not always indicate positive clinical findings, as they are often caused by other non-pathological factors like patient pose, bowel gas patterns, and clothing. For these settings, a workflow better mirroring the clinical practice, \ie robust analysis across semantic \textit{anatomical} symmetries, is needed. Using semantic anatomical symmetry to facilitate \ac{CAD} in such complex scenarios has yet to be explored. 

To bridge this gap, we propose an \ac{AASN} to effectively exploit semantic anatomical symmetry in complex imaging scenarios. Our motivation comes from the detection of pelvic fractures in emergency-room \acp{PXR}.
Pelvic fractures are among the most dangerous and lethal traumas, due to their high association with massive internal bleeding. 
Non-displaced fractures, \ie, fractures that cause no displacement of the bone structures, can be extraordinarily difficult to detect, even for experienced clinicians.
Therefore, the combination of  difficult detection coupled with extreme and highly-consequential demands on performance motivates even more progress. Using anatomical symmetry to push the performance even higher is a critical gap to fill. 




In \acp{AASN}, we employ fully convolutional Siamese networks~\cite{siamese} as the backbone of our method.
First, we exploit symmetry cues by anatomically reparameterizing the image using a powerful graph-based landmark detection~\cite{curve_gcn}. This allows us to create an anatomically-grounded warp from one side of the pelvis to the other. While previous symmetry modeling methods rely on image-based spatial alignment before encoding~\cite{mammo}, we take a different approach and perform feature alignment after encoding using a spatial transformer layer. This is motivated by the observation that image \textit{asymmetry} in \acp{PXR} can be caused by many factors, including imaging angle and patient pose. Thus, directly warping images is prone to introducing artifacts, which can alter pathological image patterns and make them harder to detect. Since image asymmetry can be semantically pathological, \ie, fractures, and non-pathological, \eg, imaging angle and patient pose, we propose a new contrastive learning component in Siamese network to optimize the deep image features being more salient corresponding to the underlying semantic asymmetries (caused by fracture). Crucially, this mitigates the impact of distracting asymmetries that may mislead the model. With a sensible embedding in place, corresponding anatomical regions are jointly decoded for fracture detection, allowing the decoder to reliably discover fracture-causing discrepancies.


In summary, our main contributions are four folds.
\begin{itemize}
    \item We present a clinically-inspired (or reader-inspired) and computationally principled framework, named \ac{AASN}, which is capable of effectively exploiting anatomical landmarks for semantic asymmetry analysis from encoded deep image features. This facilitates a high performance \ac{CAD} system of detecting both visually evident and subtle pelvic fractures in \ac{PXR}s.
    
    \item We systematically explore plausible means for fusing the image based anatomical symmetric information.  A novel Siamese feature alignment via spatial transformer layer is proposed to address the potential image distortion drawback in the prior work \cite{mammo}. 
    
    \item We describe and employ a new contrastive learning component to improve the deep image feature's representation and saliency reflected from semantically pathological asymmetries. This better disambiguates against the existing visual asymmetries caused by non-pathological reasons.
    
    \item Extensive evaluation on real clinical dataset of 2,359 \acp{PXR} from unique patients is conducted. Our results show that \ac{AASN} simultaneously increases the AUC and the average precision from $96.52\%$ to $97.71\%$ or from $94.52\%$ to $96.50\%$, respectively, compared to a strong baseline model that does not exploit symmetry or asymmetry. {\it More significantly, the pelvic fracture detection sensitivity or recall value has been boosted from $70.79\%$ to $77.87\%$ when controlling the false positive (FP) rate at $1\%$.}
\end{itemize}


\section{Related Work}

{\bf Computer-Aided Detection and Diagnosis in Medical Imaging.} In recent years, motivated by the availability of public X-ray datasets, X-ray \ac{CAD} has received extensive research attention. Many works have studied abnormality detection in \acp{CXR}~\cite{haomin_hierarchy,chestxray8,chexnet,li_thoracic_2018}. \Ac{CAD} of fractures in musculoskeletal radiographs is another well studied field~\cite{cheng2019application,hip1,pelvic_yirui}. Since many public X-ray datasets only have image-level labels, many methods formulate abnormality detection as an image classification problem and use class activation maps~\cite{zhou2016learning} for localization~\cite{chexnet,chestxray8}. While abnormalities that involve a large image area (e.g., atelectasis, cardiomegaly) may be suitable for detection via image classification, more localized abnormalities like masses and fractures are in general more difficult to detect without localization annotations. While methods avoiding such annotations have been developed~\cite{li_thoracic_2018,pelvic_yirui}, we take a different approach and use point-based localizations for annotations, which are minimally laborious and a natural fit for ill-defined fractures. Another complementary strategy to improve abnormality detection is to use anatomical and pathological knowledge and heuristics to help draw diagnostic inferences~\cite{liu2009symmetry}. This is also an approach we take, exploiting the bilateral symmetry priors of anatomical structures to push forward classification performance.




{\bf Image based Symmetric Modeling for CAD.} Because many human anatomies are left-right symmetric (e.g., brain, breast, bone), anatomical symmetry has been studied for \ac{CAD}.
The shape asymmetry of subcortical brain structures is known to be associated with Alzheimer's disease and has been measured using both analytical shape analysis~\cite{wachinger2015brainprint,konukoglu2012wesd} and machine learning techniques~\cite{Alzheimer}.  A few attempts have been explored using symmetric body parts for \ac{CAD}~\cite{stroke,mammo}. For instance,
Siamese networks~\cite{siamese} have been used to combine features of the left and right half of brain CTs for detecting strokes. A Siamese Faster-RCNN approach was also proposed to detect masses from mammograms by jointly analyzing left and right breasts~\cite{mammo}.
Yet, existing methods directly associate asymmetries in the image space with pathological abnormalities. While this assumption may hold in strictly controlled imaging scenarios, like brain CT/MRIs and mammograms, this rarely holds in \acp{PXR}, where additional asymmetry causing factors are legion, motivating the more anatomically-derived approach to symmetry that we take. 

{\bf Siamese Network and Contrastive Learning.}
Siamese networks are an effective method for contrastive learning that uses contrastive loss to embed semantically similar samples closer together and dissimilar images further away~\cite{siamese}. Local similarities have also been learned using Siamese networks~\cite{zagoruyko2015learning} and applied to achieve image matching/registration~\cite{Similarity_Matching,simonovsky2016deep}. 
The embedding learned by Siamese networks has also been applied to one-shot image recognition~\cite{koch2015siamese} and human re-identification~\cite{Similarity_Face,contrastive_face}.
Fully convolutional Siamese networks have also been proposed to produce dense and efficient sliding-window embeddings, with notable success on visual object tracking tasks~\cite{guo2017learning,bertinetto2016fully,Similarity_Tracking}.
Another popular technique for contrastive learning is triplet networks~\cite{triplet}. We also use Siamese networks to learn embeddings; however, we propose a process to learn embeddings that are invariant to spurious asymmetries, while being sensitive to pathology-inducing ones. 




\section{Method}


\subsection{Problem Setup}



\label{sec:transformation}
\begin{figure}[t]
   \begin{center}
      \includegraphics[width=0.8\linewidth]{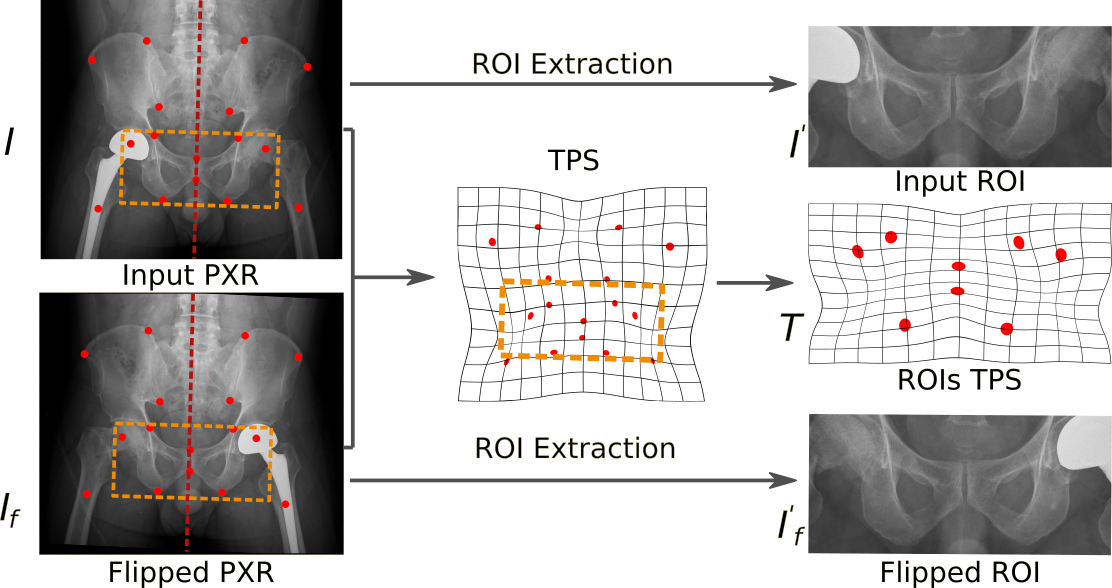}
   \end{center}
   \caption{Illustration of ROI and warp generation steps.}
   \label{fig:transformation}
\end{figure}

Given a \ac{PXR}, denoted as $I$, we aim to detect sites of anterior pelvic fractures. Following the widely adopted approach by \ac{CAD} methods~\cite{li_thoracic_2018,wang2018chestnet,chestxray8}, our model produces image-level binary classifications of fracture and heatmaps as fracture localization. Using heatmaps to represent localization (instead of bounding box or segmentation) stems from the inherent ambiguity in the definition of instance and boundary of pathological abnormalities in medical images. For instance, a fracture can be comminuted, \ie bone breaking into multiple pieces, resulting in ambiguity in defining the number of fractures. Our model takes a cost-effective and flexible annotation format, a point at the center of each fracture site, allowing ambiguous fracture conditions to be flexibly represented as one point or multiple points.
We dilate the annotation points by an empirically-defined radius (2 cm in our experiment) to produce a mask for the \ac{PXR}, which is the training target of our method, denoted as $M$. In this way, we execute heatmap regression, similar to landmark detection~\cite{xu2018less}, except for center-points of abnormalities with ambiguous extents. 


\subsection{Anatomy-Grounded Symmetric Warping}
Given the input \ac{PXR} image, our method first produces \ac{ROI} of the anterior pelvis and anatomically-grounded warp to reveal the bilateral symmetry of the anatomy. The steps of \ac{ROI} and warp generation are illustrated in Figure~\ref{fig:transformation}. First, a powerful graph-based landmark detection~\cite{li2020structured} is applied to detect 16 skeletal landmarks, including 7 pairs of bilateral symmetric landmarks and 2 points on pubic symphysis. From the landmarks, a line of bilateral symmetry is regressed, and the image is flipped with respect to it. Since we focus on detecting anterior pelvic fractures, where the dangers of massive bleeding is high and fractures are hard to detect, we extract \acp{ROI} of the anterior pelvis from the two images as a bounding box of landmarks on the pubis and ischium, which are referred as $I$ and $I_f$. A pixel-to-pixel warp from $I_f$ to $I$ is generated from the corresponding landmarks in $I_f$ and $I$ using the classic \ac{TPS} warp~\cite{thin_plate_spline}, denoted as $T$. Note, the warp $T$ is not directly used to align the images. Instead, it is used in our Siamese network via a spatial transformer layer to align the features. 



\subsection{Anatomy-Aware Siamese Network}
\label{sec:AASN}

\begin{figure}[t]
   \begin{center}
      \includegraphics[width=\textwidth]{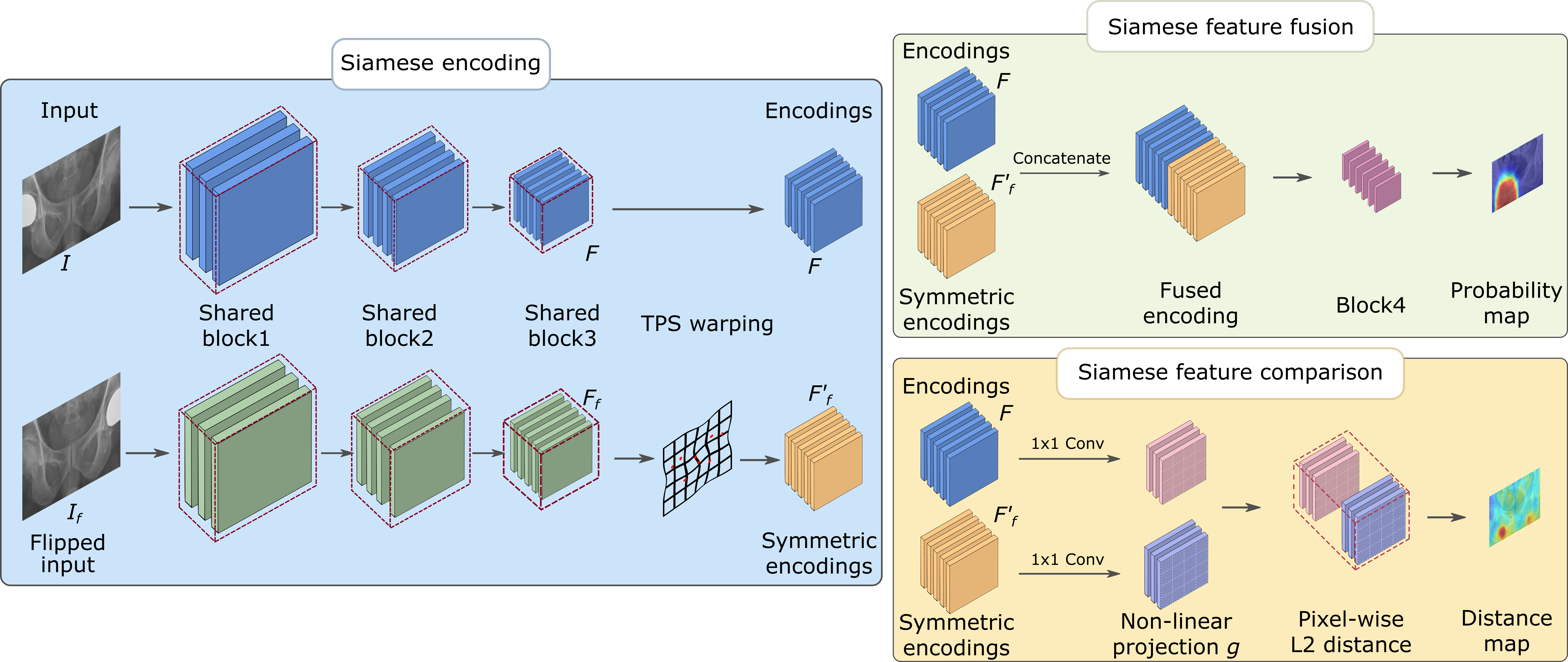}
   \end{center}
   \caption{System overview of the  proposed \ac{AASN}. The Siamese encoding module takes two pre-processed ROIs as input and encodes them using dense blocks with shared weights. After warping and alignment, the encoded feature maps are further processed by a Siamese feature fusion module and a Siamese contrastive learning module to produce a fracture probability map and a feature distance map, respectively.}
   \label{fig:framework}
\end{figure}


The architecture of \ac{AASN} is shown in Figure~\ref{fig:framework}. \ac{AASN} contains a fully convolutional Siamese network with a DenseNet-121~\cite{huang2016densely} backbone. The dense blocks are split into two parts, an encoding part and a decoding part. It is worth noting that \ac{AASN} allows the backbone network to be split flexibly at any block. For our application, we split at a middle level after the 3rd dense block, where the features are deep enough to encode the local skeletal pattern, but has not been pooled too heavily so that the textual information of small fractures is lost.

The encoding layers follow a Siamese structure, with two streams of weight-shared encoding layers taking the two images $I$ and $I_f$ as inputs. The encoder outputs, denoted as $F$ and $F_f$, provide feature representations of the original image and the flipped image, respectively. The spatial alignment transform $T$ is applied on $F_f$, resulting in $F'_f$, making corresponding pixels in $F$ and $F'_f$ represent corresponding anatomies. The two aligned feature maps are then fused and decoded to produce a fracture probability map, denoted as $Y$. Details of feature map fusion and decoding will be described in Sec.~\ref{sec:feature_fusion}. We produce the probability heatmap as fracture detection result to alert the clinician the presence of a fracture and also to guide his or her attention (as shown in Figure~\ref{fig:detection_results}). Since pelvis fractures can be very difficult to detect, even when there is a known fracture, this localization is a key feature over-and-above image-level predictions. 

The model is trained using two losses. The first loss is the pixel-wise \ac{BCE} between the predicted heatmap $Y$ and the ground truth $M$, denoted as $L_b$. The second loss is the pixel-wise contrastive loss between the two feature maps, $F$ and $F'_f$, denoted as $L_c$. Details of the contrastive loss will be discussed in Sec.~\ref{sec:embedding_learning}. The total loss can be written as
\begin{align}
\label{equ:total_loss}
    L = L_b + \lambda L_c,
\end{align}
where $\lambda$ is a weight balancing the two losses.

\subsection{Siamese Feature Fusion}
\label{sec:feature_fusion}

\begin{figure}[t]
   \begin{center}
      \includegraphics[width=0.8\linewidth]{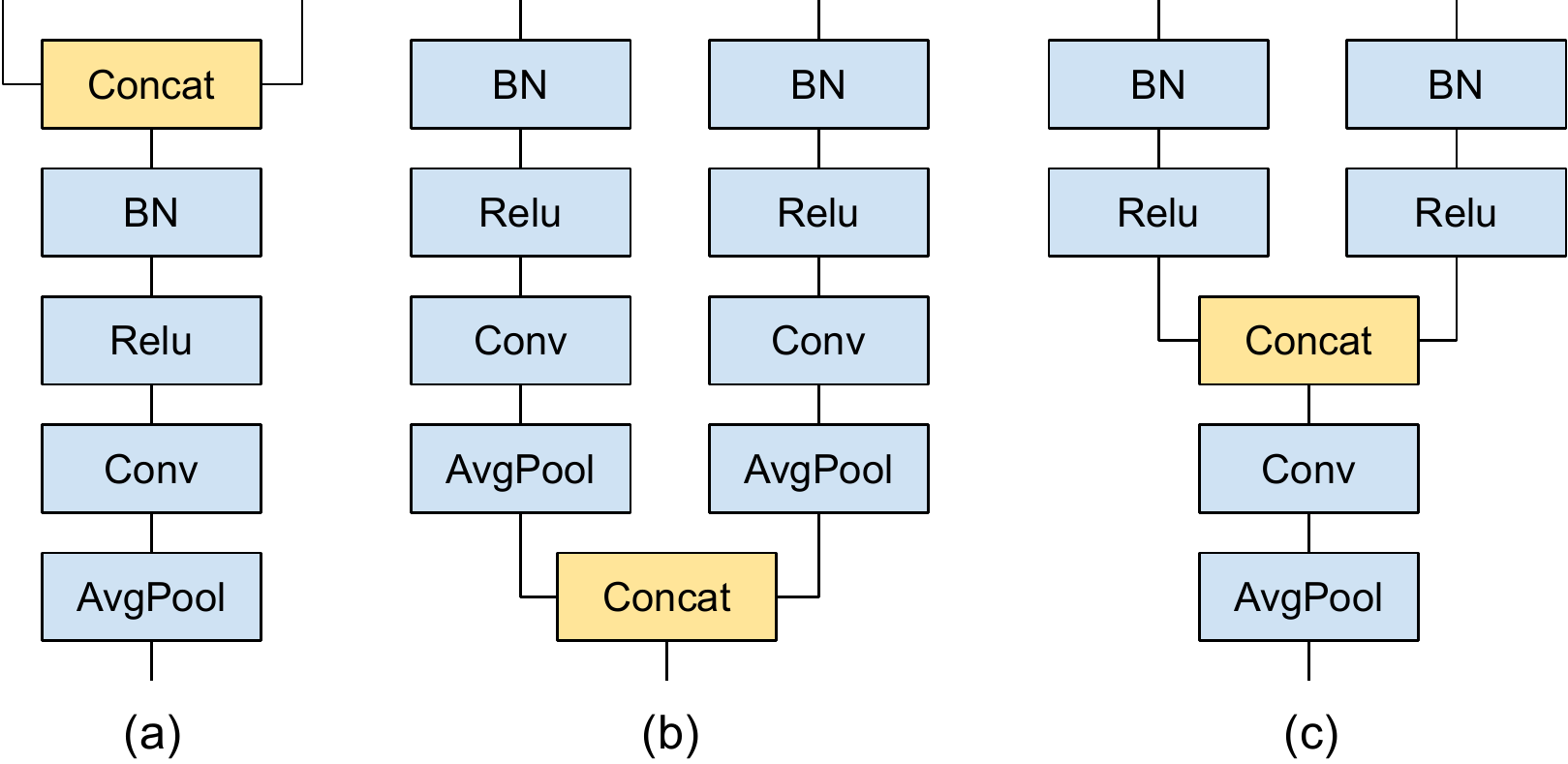}
   \end{center}
   \caption{Transition layer modification options for feature map fusion. (a) Feature map fusion before transition. (b) Feature map fusion after transition. (c) Feature map fusion inside transition}
   \label{fig:feature_fusion}
\end{figure}

The purpose of encoding the flipped image is to provide a reference of the symmetric counterpart, $F_{f}$, which can be incorporated with the feature $F$ to facilitate fracture detection. To provide a meaningful reference, $F_f$ needs to be spatially aligned with $F$, so that features with the same index/coordinate in the two feature maps encode the same, but symmetric, anatomies of the patient. Previous methods have aligned the bilateral images $I$ and $I_f$ directly before encoding~\cite{mammo}. However, when large imaging angle and patient pose variations are present, image alignment is prone to introducing artifacts, which can increase the difficulty of fracture detection. Therefore, instead of aligning the bilateral images directly, we apply a spatial transformer layer on the feature map $F_f$ to align it with $F$, resulting in $F'_f$. The aligned feature maps $F$ and $F'_f$ are fused to produce a bilaterally combined feature map, where every feature vector encodes the visual patterns from symmetrical anatomies. This allows the decoder to directly incorporate symmetry analysis into fracture detection.



We fuse the feature maps by concatenation. Implementation of the concatenation involves modification to the transition module between the dense blocks, where multiple options exist, including concatenation before, after, or inside the transition module (as shown in Figure~\ref{fig:feature_fusion}).
A transition module in DenseNet consists of sequential BatchNorm, ReLU, Conv and AvgPool operations. We perform the concatenation inside the transition module after the ReLU layer, because it causes minimal structural changes to the DenseNet model. Specifically, the only layer affected in the DenseNet is the $1\times 1$ Conv layer after concatenation, whose input channels are doubled. All other layers remain the same, allowing us to leverage the ImageNet pre-trained weights.



\subsection{Siamese Contrastive Learning}
\label{sec:embedding_learning}


While the above feature fusion provides a principled way to perform symmetric analysis, further advancements can be made. We are motivated by a key insight that image asymmetry can be caused by pathological abnormalities, \ie fracture, or spurious non-pathological factors, \eg soft tissue shadows, bowel gas patterns, clothing and foreign bodies. These non-pathological factors can be visually confusing, causing false positives. We aim to optimize the deep features to be more salient to the semantically pathological asymmetries, while mitigating the impact of distracting non-pathological asymmetries.
To this end, our model employs a new constrastive learning component to minimize the pixel-wise distance between $F$ and $F'_f$ in areas without fracture, making the features insensitive to non-semantic asymmetries and thus less prone to false positives. On the other hand, our contrastive learning component encourages larger distance between $F$ and $F'_f$ in areas with fractures, making the features more sensitive to semantic asymmetries.

The above idea is implemented using pixel-wise margin loss between $F$ and $F'_f$ after a non-linear projection $g$:
\begin{equation}
  L_c = \sum_{\boldsymbol{x}}
    \begin{cases}
      \| g(F(\boldsymbol{x})) - g(F'_f(\boldsymbol{x})) \|^2  & \text{if } \boldsymbol{x} \notin \hat{M}\\
      \max (0, m - \| g(F(\boldsymbol{x})) - g(F'_f(\boldsymbol{x})) \|^2) & \text{if } \boldsymbol{x} \in \hat{M}\\
    \end{cases},
\end{equation}
where $\boldsymbol{x}$ denotes the pixel coordinate, $\hat{M}$ denotes the mask indicating areas affected by fractures, and $m$ is a margin governing the dissimilarity of semantic asymmetries. 
The mask $\hat{M}$\ is calculated as $\hat{M} = M \cup T \circ M_f$, where $T \circ M_f$ is flipped and warped $M$.

{\color{black}
We employ a non-linear projection $g$ to transform the feature before calculating the distance, which improves the quality of the learned feature $F$, $F_f'$. In our experiment, the non-linear projection consists of a linear layer followed by BatchNorm and ReLU. We posit that directly performing contrast learning on features used for fracture detection could induce information loss and limit the modeling power. For example, bone curvature asymmetries in X-ray images are often non-pathological (e.g., caused by pose). However, they also provide visual cues to detect certain types of fractures. Using the non-linear projection, such useful information can be excluded from the contrastive learning so that they are preserved in the feature for the downstream fracture detection task. 

While the margin loss has been adopted for \ac{CAD} in a previous method~\cite{Alzheimer}, it was employed as a metric learning tool to learn a distance metric that directly represent the image asymmetry. We stress that our targeted \ac{CAD} is more complex and clinically relevant, where image asymmetry can be semantically non-pathological (caused by pose, imaging condition and etc.) but we are only interested in detecting the pathological (fracture-caused) asymmetries. We employ the margin loss in our contrastive learning component to learn features with optimal properties. For this purpose, extra measures are taken in our method, including 1) conducting multi-task training with the margin loss calculated on a middle level feature, and 2) employing a non-linear projection head to transform the feature before calculating the margin loss.
}

\section{Experiments}

We demonstrate that our proposed AASN can significantly improve the performance in pelvic fracture detection by exploiting the semantic symmetry of anatomies. We focus on detecting fractures on the anterior pelvis including pubis and ischium, an anatomically symmetric region with high rate of diagnostic errors and life-threatening complications in the clinical practice. 

\subsection{Experimental Settings}
\textbf{Dataset: }
We evaluate \ac{AASN} on a real-world clinical dataset collected from the Picture Archiving and Communication System (PACS) of a hospital's trauma emergency department. The images have a large variation in the imaging conditions, including viewing angle, patient pose and foreign bodies shown in the images. Fracture sites in these images are labeled by experienced clinicians, combining multiple sources of information for confirmation, including clinical records and computed tomography scans. The annotations are provided in the form of points, due to inherent ambiguity in defining fracture as object. In total, there are $2\,359$ PXRs, and $759$ of them have at least one anterior pelvic fracture site. All our experiments are conducted with five-fold cross-validation with a $70\%/10\%/20\%$ training, validation, and testing split, respectively.


\textbf{Implementation Details: }
The ROIs of the anterior pelvis are resized to $256 \times 512$ and stacked to a 3-channel pseudo-color image. We produce the supervision mask for the heatmap prediction branch by dilating the annotation points to circle masks with a radius of $50$ (about 2 cm). We implement all models using PyTorch~\cite{pytorch}. Severe over-fitting is observed when training the networks from scratch, so we initialize them with ImageNet pre-trained weights. We emperically select DenseNet-121 as the backbone which yields the best performance comparing to other ResNet and DenseNet settings. All models are optimized by Adam~\cite{adamopt} with a learning rate of $10^{-5}$. For the pixel-wise contrastive loss, we use the hyperparameter $m=0.5$ as the margin, and $\lambda = 0.5$ to balance the total loss.

\textbf{Evaluation Metrics: }
We first assess the model's performance as an image-level classsifier, which is a widely adopted evaluation approach for CAD systems~\cite{li_thoracic_2018,wang2018chestnet,chestxray8}. The image-level abnormality reporting is of utmost importance in clinical workflow because it directly affects the clinical decision. We take the maximum value of the output heatmap as the classification output, and use Area under ROC Curve (AUC) and Average Precision (AP) to evaluate the classification performance.

\begin{figure}[t]
   \begin{center}
      \includegraphics[width=\linewidth]{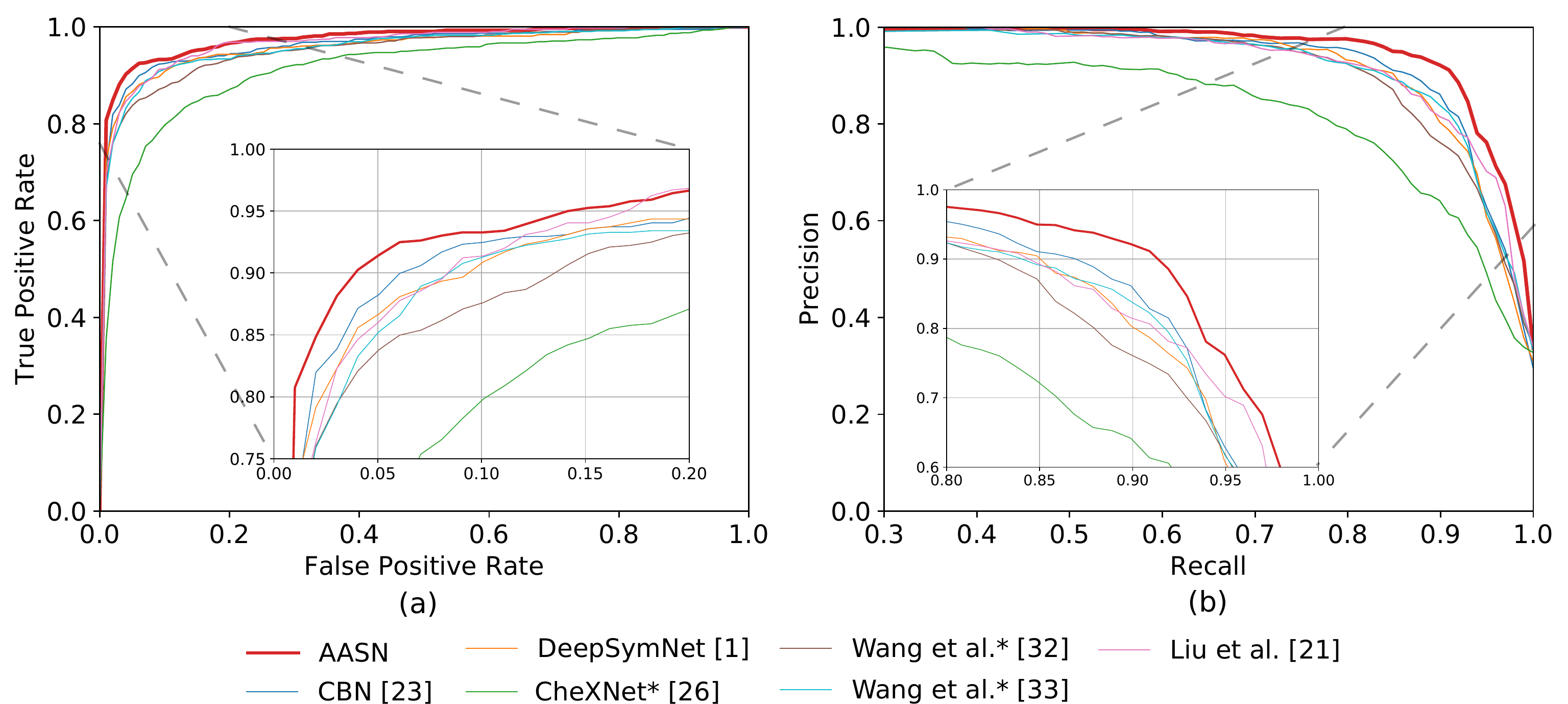}
   \end{center}
   \caption{Comparison of ROC curve and PR curve to the baselines. (a) is the ROC curve and (b) is the PR curve. *Methods trained using image-level labels.}
   \label{fig:pr_roc}
\end{figure}

{
\color{black}
We also evaluate the model's fracture localization performance. Since our model produces heatmaps as fracture localization, standard object detection metrics do not apply. A modified free-response ROC (FROC) is reported to measure localization performance. Specifically, unlike FROC, where object recall is reported with the number of false positives per image, we report fracture recall with the ratio of false positive area per image. A fracture is considered recalled if the heatmap activation value at its location is above the threshold. Areas with $>$2 cm away from all fracture annotation points are considered negative, on which the false positive ratio is calculated. Areas within 2 cm from any annotation point is considered as ambiguous extents of the fracture. Since both positive and negative responses in these ambiguous areas are clinically acceptable, they are excluded from the modified FROC calculation.
}

\textbf{Compared Methods: }
We first compare \ac{AASN} with three state-of-the-art CAD methods, \ie, ChexNet~\cite{chexnet}, Wang \etal{}~\cite{chestxray8}, and Wang \etal{}~\cite{pelvic_yirui}, all using image-level labels for training. They classify abnormality at image-level, and output heatmaps for localization visualization. ChexNet~\cite{chexnet} employs a global average pooling followed by a fully connected layer to produce the final prediction. Wang \etal{}~\cite{chestxray8} uses Log-Sum-Exp (LSE) pooling. Wang \etal{}~\cite{pelvic_yirui} employs a two-stage classification mechanism, and reports the state-of-the-art performance on hip/pelvic fracture classification.

We also compare with three methods modeling symmetry for CAD, \ie, Liu \etal{}~\cite{Alzheimer}, CBN~\cite{mammo} and DeepSymNet~\cite{stroke}. All three methods perform alignment on the flipped image.  
Liu \etal{}~\cite{Alzheimer} performs metric learning to learn a distance metric between symmetric body parts and uses it directly as an indicator of abnormalities. DeepSymNet~\cite{stroke} and CBN~\cite{mammo} fuse the Siamese encodings for abnormality detection, using subtraction and concatenation with gating, respectively.
All evaluated methods use DenseNet-121 backbone, trained using the same experiment setting and tested with five-fold cross validation.

\setlength{\tabcolsep}{2mm}
\begin{table}[t]
\centering
\caption{Fracture classification and localization performance comparison with state-of-the-art models. Classifier AUC and AP are reported for classification performance. Fracture recalls at given false positive ratio are reported for localization performance. *Methods trained using image-level labels. Localization performance are not evaluated on these methods.}
\vspace{1em}
\begin{tabular}{lcccc}
\toprule
\multirow{2}{*}{Method}  & \multicolumn{2}{c}{Classification} & \multicolumn{2}{c}{Localization} \\ \cmidrule(lr){2-3} \cmidrule(lr){4-5}
   & AUC & AP & Recall\textsubscript{FP=1\%} & Recall\textsubscript{FP=10\%}  \\ \midrule
CheXNet*~\cite{chexnet}           & 93.42\%         & 86.33\%  & - & - \\
Wang \etal{}*~\cite{chestxray8}   & 95.43\%         & 93.31\%  & - & - \\
Wang \etal{}*~\cite{pelvic_yirui} & 96.06\%         & 93.90\%  & - & - \\ \midrule
Liu \etal{}~\cite{Alzheimer}      & 96.84\%         & 94.29\%  & 2.78\%  &  24.19\% \\
DeepSymNet~\cite{stroke}          & 96.29\%         & 94.45\%  & 69.66\% &  90.07\% \\
CBN~\cite{mammo}                  & 97.00\%         & 94.92\%  & 73.93\% &  90.90\% \\
\ac{AASN}                         &\textbf{97.71\%} & \textbf{96.50\%}  & \textbf{77.87\%} & \textbf{92.71\%}       \\ \bottomrule
\end{tabular}
\label{tab:existing_method_comparison}
\end{table}

\subsection{Classification Performance}

Evaluation metrics of fracture classification performance are summarized in Table~\ref{tab:existing_method_comparison}. ROC and PR curves are shown in Figure~\ref{fig:pr_roc}. The methods trained using only image-level labels result in overall lower performance than methods trained using fracture sites annotations. AASN outperforms all other methods, including the ones using symmetry and fracture site annotations, with substantial margins in all evaluation metrics. The improvements are also reflected in the ROC and PR curves~Figure~\ref{fig:pr_roc}. Specifically, comparing to the 2nd highest values among all methods, AASN improves AUC and AP by 0.71\% and 1.58\%, from 97.00\% and 94.92\% to 97.71\% and 96.50\%, respectively. We stress that in this high AUC and AP range (\ie above 95\%), the improvements brought by AASN are significant. For instance, when recall is increased from 95\% to 96\%, the number of missed fractures are reduced by 20\%.

\begin{figure}[t]
   \begin{center}
      \includegraphics[width=\linewidth]{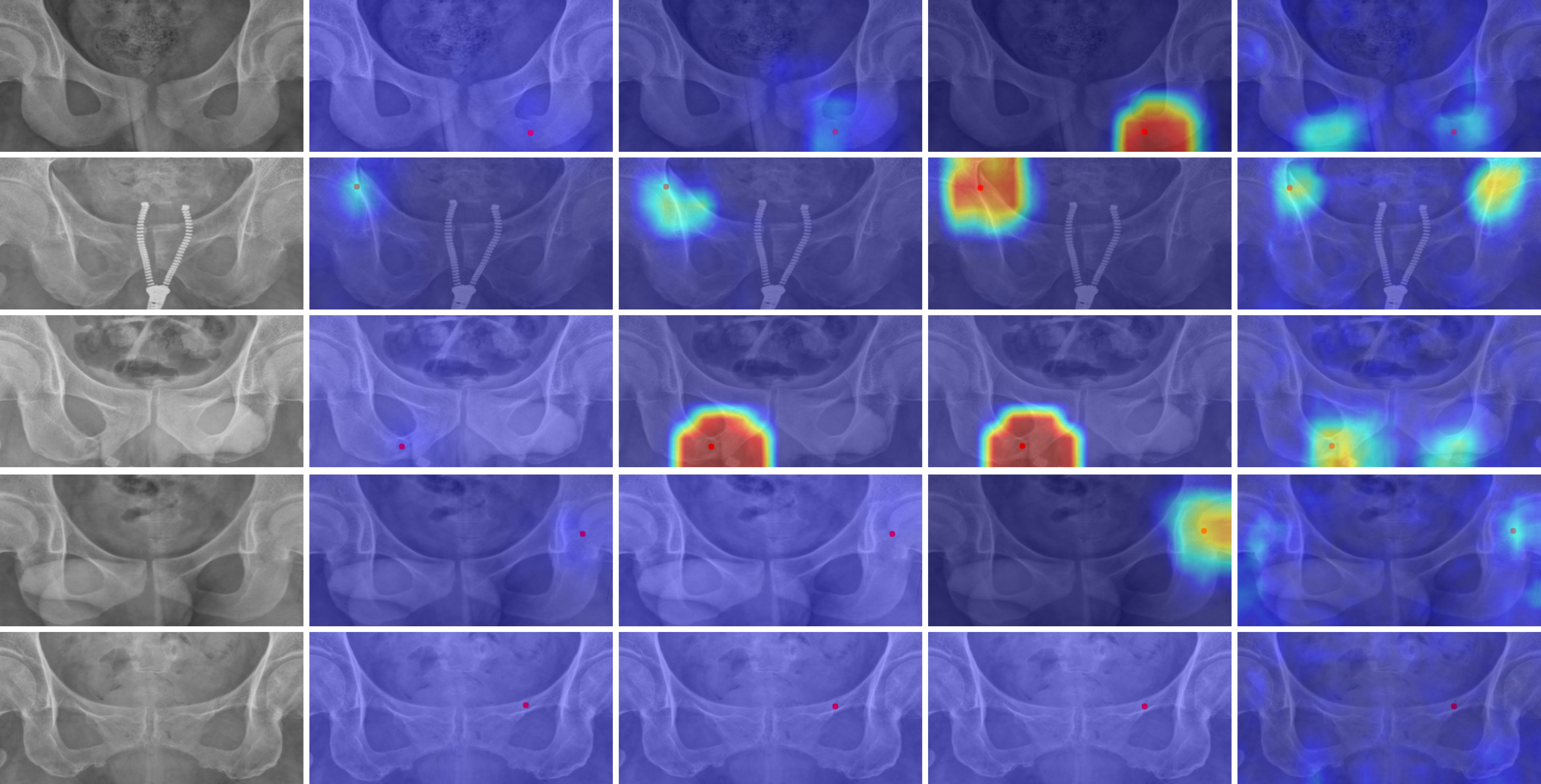}
      \newline
      \makebox[0.19\linewidth]{(a)}
      \makebox[0.19\linewidth]{(b)}
      \makebox[0.19\linewidth]{(c)}
      \makebox[0.19\linewidth]{(d)}
      \makebox[0.19\linewidth]{(e)}
   \end{center}
   \caption{Prediction results for different models. (a) pubis ROI in the PXR. Fracture probability heatmaps produced by (b) Vanilla DenseNet-121~\cite{densenet}, (c) CBN~\cite{mammo} and (d) AASN. (e) the distance map between siamese feature in AASN. The last row shows an example of failed cases.}
   \label{fig:detection_results}
\end{figure}

Figure~\ref{fig:detection_results} provides visualizations of fracture heatmaps produced using different methods. Non-displaced fractures that do not cause bone structures to be largely disrupted are visually ambiguous and often missed by the vanilla DenseNet-121 without considering symmetry. Comparison between the fracture site and its symmetric bone reveals that the suspicious pattern only occurs on one side and is likely to be fracture. 
This intuition is in line with the results, \ie, by incorporating symmetric features, some of the ambiguous fractures can be detected. 
By employing the feature comparison module, AASN is able to detect more fracture, hypothetically owing to the better feature characteristics learned via feature comparison.

\subsection{Localization Performance}

We also evaluate AASN's fracture localization performance. The three symmetry modeling baselines and our four ablation study methods are also evaluated for comparison.
As summarized in Table~\ref{tab:existing_method_comparison}, AASN achieves the best fracture site recall among all evaluated methods, resulting in Recall\textsubscript{FP=1\%}=77.87\% and Recall\textsubscript{FP=10\%}=92.71\%, respectively. It outperforms baseline methods by substantial margins.

Among the baseline methods, directly using learned distance metric as an indicator of fracture (Liu \etal{}~\cite{Alzheimer}) results in the lowest localization performance, because the image asymmetry indicated by distance metric can be caused by other non-pathological factors than fractures. The comparison justifies the importance of our proposed contrastive learning component, which \textit{exploits image asymmetry to optimize deep feature for downstream fracture detection}, instead of directly using it as a fracture indicator. CBN~\cite{mammo} achieves the best performance among the three baselines, hypothetically owing to the Siamese feature fusion. With our feature alignment and contrastive learning components, AASN significantly improves fracture site Recall\textsubscript{FP=1\%} over CBN~\cite{mammo} by 3.94\%.

\subsection{Ablation Study}

\renewcommand{\arraystretch}{0.9}
\setlength{\tabcolsep}{2mm}
\begin{table}[t]
\centering
\caption{Ablation study of AASN. The baseline model is vanilla DenseNet121 trained without the symmetry modeling components. ``FF'' indicates using feature fusion. ``FA'' indicates using feature alignment (otherwise image alignment is used). ``CL'' indicates using contrastive learning. ``no. proj.'' indicates that the contrastive learning is performed without the non-linear projection head.}
\begin{tabular}{lllcccc}
\toprule
FF & FA & CL & AUC              & AP   & Recall\textsubscript{FP=1\%} & Recall\textsubscript{FP=10\%} \\ \midrule
   &    &    & 96.52\%          & 94.52\%  & 70.79\% & 89.46\%  \vspace{1mm}        \\
 \multirow{2}{*}{\checkmark} &            &            & 96.93\%  & 94.77\% & 73.22\% & 89.93\%  \vspace{-1mm}\\ 
           &            &            & \scriptsize{\color{blue} (+0.41\%)} & \scriptsize{\color{blue} (+0.25\%)} & \scriptsize{\color{blue} (+2.43\%)} & \scriptsize{\color{blue} (+0.47\%)}  \vspace{1mm}\\ 
\multirow{2}{*}{\checkmark} & \multirow{2}{*}{\checkmark} &            & 97.20\%         & 95.68\% & 76.68\% & 91.51\%    \vspace{-1mm}     \\
           &            &            & \scriptsize{\color{blue} (+0.68\%)} & \scriptsize{\color{blue} (+1.16\%)} & \scriptsize{\color{blue} (+5.89\%)} & \scriptsize{\color{blue} (+2.05\%)} \vspace{1mm}\\ 
\multirow{2}{*}{\checkmark} &            & \multirow{2}{*}{\checkmark} & 97.46\%         & 95.36\% & 76.27\% & 91.09\%   \vspace{-1mm}      \\ 
           &            &            & \scriptsize{\color{blue} (+0.94\%)} & \scriptsize{\color{blue} (+0.84\%)} & \scriptsize{\color{blue} (+5.48\%)} & \scriptsize{\color{blue} (+1.63\%)} \vspace{1mm}\\ \midrule
\multirow{2}{*}{\checkmark} & \multirow{2}{*}{\checkmark} & \multirow{2}{*}{\checkmark \textsubscript{no proj.}} & 97.31\% & 96.15\%  & 77.26\% &92.70\%   \vspace{-1mm}      \\
           &            &            & \scriptsize{\color{blue} (+0.79\%)} & \scriptsize{\color{blue} (+1.63\%)} & \scriptsize{\color{blue} (+6.47\%)} & \scriptsize{\color{blue} (+3.24\%)} \vspace{1mm}\\ 
\multirow{2}{*}{\checkmark} & \multirow{2}{*}{\checkmark} & \multirow{2}{*}{\checkmark} & \bf 97.71\%  & \textbf{96.50\%} & \textbf{77.87\%} & \textbf{92.71\%} \vspace{-1mm} \\ 
           &            &            & \scriptsize{\color{blue} (+1.19\%)} & \scriptsize{\color{blue} (+1.98\%)} & \scriptsize{\color{blue} (+7.08\%)} & \scriptsize{\color{blue} (+3.25\%)} \\ \bottomrule
\end{tabular}
\label{tab:ablition}
\end{table}


We conduct ablation study of AASN to analyze the contributions of its novel components, summarized in Table \ref{tab:ablition}. The components include: 1) Symmetric feature fusion (referred to as \textit{FF}), 2) Feature alignment (referred to as \textit{FA}) and 3) Feature contrastive learning (referred to as \textit{CL}). We add these components individually to the Vanilla DenseNet-121 to analyze their effects. We also analyze the effect of the non-linear projection head $g$ by evaluating a variant of constrastive learning without it.


\textbf{Symmetric Feature Fusion:}
The effect of feature fusion is reflected in the comparisons: baseline vs. \textit{FF} and baseline vs. \textit{FF-FA}. Both \textit{FF} and \textit{FF-FA} employ symmetric feature fusion and are able to outperform \textit{Vanilla}, although by a different margin due to the different alignment methods used. In particular, \textit{FF-FA} significantly improves the Recall\textsubscript{FP=1\%} by 5.89\%. 
These improvements are hypothetically owing to the incorporation of the visual patterns from symmetric body parts, which provides reference for differentiating visually ambiguous fractures.

\textbf{Feature Alignment:}
The effect of feature warping and alignment is reflected in the comparisons: \textit{FF} vs. \textit{FF-FA} and \textit{FF-CL} vs. \textit{FF-FA-CL}. The ablation study shows that, by using the feature warping and alignment, the performances of both \textit{FF} and \textit{FF-CL} are both significantly improved. In particular, the Recall\textsubscript{FP=1\%} are improved by $3.46\%$ and $1.60\%$ in \textit{FF-FA} and \textit{FF-FA-CL}, respectively. 
It's also demonstrated that the contributions of feature warping and alignment are consistent with and without Siamese feature comparison.
We posit that the performance improvements are owing to the preservation of the original image pattern by performing warping and alignment at the feature level.

\textbf{Contrastive Learning:}
The effect of Siamese feature comparison is reflected in the comparisons: \textit{FF} vs. \textit{FF-CL} and \textit{FF-FA} vs. \textit{FF-FA-CL}. The ablation study shows measurable contribution of the Siamese feature comparison module. By using Siamese feature fusion, \textit{FF} and \textit{FF-FA} already show significant improvements comparing to the baseline.
By adding Siamese feature comparison to \textit{FF} and \textit{FF-FA}, Recall\textsubscript{FP=1\%} are improved by $3.05\%$ and $1.19\%$, respectively. The improvements are in line with our motivation and hypothesis that by maximizing/minimizing Siamese feature distances on areas with/without fractures, the network can learn features that are more sensitive to fractures and less sensitive to other distracting factors. Comparing to the AASN directly performing constrastive learning on the symmetric feature (\textit{no. proj.}), employing the non-linear projection head leads further improves the Recall\textsubscript{FP=1\%} by 0.61\%.

\section{Conclusion}
In this paper, we systematically and thoroughly study exploiting the anatomical symmetry prior knowledge to facilitate \ac{CAD}, in particular anterior pelvic fracture detection in \ac{PXR}. We introduce a deep neural network technique, termed Anatomical-Aware Siamese Network, to incorporate semantic symmetry analysis into abnormality (\ie fracture) detection. Through comprehensive ablation study, we demonstrate that: 1) Employing symmetric feature fusion can effectively exploit symmetrical information to facilitate fracture detection. 2) Performing spatial alignment at the feature level for symmetric feature fusion leads to substantial performance gain. 3) Using contrastive learning, the Siamese encoder is able to learn more sensible embedding, leading to further performance improvement. By comparing with the state-of-the-art methods, including latest ones modeling symmetry, we demonstrate the \ac{AASN} is by far the most effective method exploiting symmetry and reports substantially improved performances on both classification and localization tasks. 

\clearpage
%
%
\bibliographystyle{splncs04}
\bibliography{reference}
\end{document}